\documentclass[10pt,twocolumn,letterpaper]{article}

\usepackage{iccv,flushend}
\usepackage{times}
\usepackage{epsfig}
\usepackage{graphicx}
\usepackage{amsmath,amssymb} 
\usepackage{amsthm}
\usepackage{overpic}
\newtheorem{theorem}{Theorem}
\newtheorem{lemma}{Lemma}
\usepackage[table]{xcolor}
\newcommand{\specialcell}[2][c]{\begin{tabular}[#1]{@{}c@{}}#2\end{tabular}} 

\usepackage[breaklinks=true,bookmarks=false]{hyperref}

\iccvfinalcopy 

\def\iccvPaperID{****} 

\ificcvfinal\fi

\begin{document}

\title{ALFA -- Leveraging All Levels of Feature Abstraction for Enhancing the \\Generalization of Histopathology Image Classification Across Unseen Hospitals}

\author{Milad Sikaroudi$^1$, Maryam Hosseini$^1$, Shahryar Rahnamayan$^2$, H.R. Tizhoosh$^{1,3}$\\ \\
$^1$Kimia Lab, University of Waterloo, ON, Canada\\
$^2$ Department of Engineering, Brock University, ON, Canada\\
$^3$Rhazes Lab, Mayo Clinic, Rochester, MN, USA\\
}

\maketitle
\ificcvfinal\thispagestyle{empty}\fi

\begin{abstract}
   We propose an exhaustive methodology that leverages all levels of feature abstraction, targeting an enhancement in the generalizability of image classification to unobserved hospitals. Our approach incorporates augmentation-based self-supervision with common distribution shifts in histopathology scenarios serving as the pretext task. This enables us to derive invariant features from training images without relying on training labels, thereby covering different abstraction levels. Moving onto the subsequent abstraction level, we employ a domain alignment module to facilitate further extraction of invariant features across varying training hospitals. To represent the highly specific features of participating hospitals, an encoder is trained to classify hospital labels, independent of their diagnostic labels.  The features from each of these encoders are subsequently disentangled to minimize redundancy and segregate the features. This representation, which spans a broad spectrum of semantic information, enables the development of a model demonstrating increased robustness to unseen images from disparate distributions. Experimental results from the PACS dataset (a domain generalization benchmark), a synthetic dataset created by applying histopathology-specific jitters to the MHIST dataset (defining different domains with varied distribution shifts), and a Renal Cell Carcinoma dataset derived from four image repositories from TCGA, collectively indicate that our proposed model is adept at managing varying levels of image granularity. Thus, it shows improved generalizability when faced with new, out-of-distribution hospital images.
\end{abstract}

\section{Motivation}

In Domain Generalization (DG), domain alignment techniques are popular \cite{tzeng2014deep,long2015learning,dou2019domain}, as they aim to minimize differences among source domains to learn domain-invariant features that can withstand unforeseen shifts in the target domain \cite{sun2016return}. 

Ignoring domain-specific information in favor of domain-invariant features may not always lead to the best generalization performance, as noted by Mancini et al. \cite{mancini2018best} and Shankar et al. \cite{shankar2018generalizing}. Bui et al. \cite{bui2021exploiting} proposed the mDSDI method and provided mathematical proof for that.

In the field of histopathology, various domains are often represented by different hospitals. In such settings, biases or distribution shifts can emerge due to the process of sample collection, slide preparation, or interpretation \cite{esteva2017dermatologist,gurcan2009histopathological}. These shifts can potentially create a mismatch between training and testing data. Recognizing these potential biases is paramount for ensuring precise and dependable outcomes in computational pathology \cite{madabhushi2016image}. 

\begin{figure}[!t]
    \centering
    \includegraphics[trim= 10 15 90 15, clip]{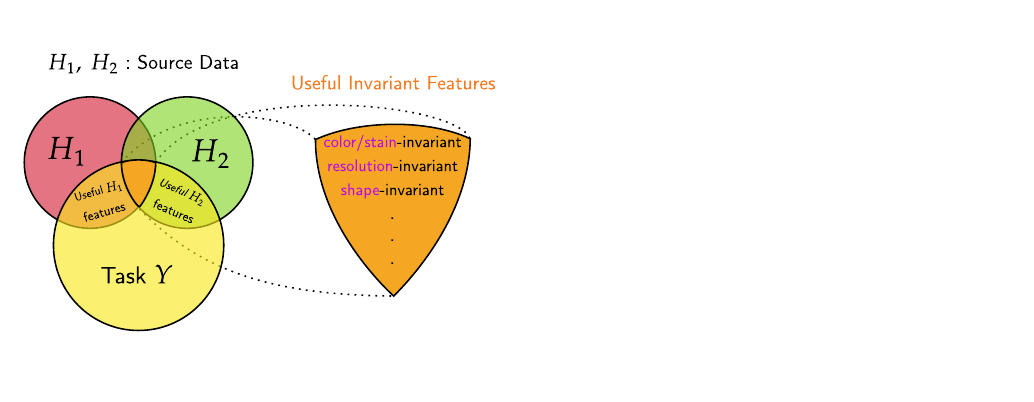}
    \caption{The Venn diagram delineates the feature space of the source hospitals ($H_1$ and $H_2$). The yellow area demarcates the label space employed in the classification task. The shared space between feature and label spaces underscores the features that are conducive to executing tasks within the label space.}
    \label{fig:venn}
\end{figure}

Figure \ref{fig:venn} illustrates that, apart from the invariant features, there exists a set of unique features specific to each hospital. These features are of significant importance as they can establish a mapping from the feature space to the label space or perform the task $Y$.

\begin{figure}
    \centering
    \includegraphics[trim= 0cm 0cm 10.5cm 0cm, clip, width= 0.3\textwidth]{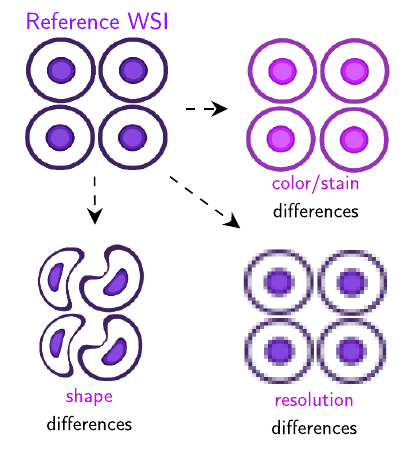}
    \caption{Different distribution shifts happening in digital pathology setups. }
    \label{fig:differences}
\end{figure}

Although acknowledging these biases may lessen distribution shifts, their total elimination, especially in computational pathology, remains unattainable \cite{beck2011systematic}. Several factors, such as the absence of standardization across pathology labs, the need for diverse inclusion, variable slide preparation guidelines, and inter-observer variability \cite{beck2011systematic}, result in \emph{visual} and \emph{histomorphological} differences in histopathology slides. Notably, different races or sexes might display distinct histomorphologic patterns \cite{rawla2019epidemiology}, thereby contributing to distribution shifts between training and testing data, as demonstrated in Fig. \ref{fig:differences}.

As a result, valuable invariant features such as (1) color/stain, (2) resolution, and (3) shape, among others, can be identified in digital pathology setups as shown in Fig. \ref{fig:venn}. The distribution shifts across various hospitals or domains are typically unknown beforehand. However, self-supervision that mimics these differences (through augmentation) can help the model learn representations invariant to these variations. 

Self-supervision nudges the model to derive features useful for predicting the data itself, independent of any prevalent hard labels \cite{pathak2016context,chen2020simple}. Consequently, this can lead to the capture of more fundamental perceptual information like edges, corners, and textures in digital images \cite{caron2020unsupervised}. The amalgamation of these self-supervised representations with other invariant and specific ones allows for a spectrum of representations, covering all levels of \emph{feature abstraction}.

This chapter proposes a new method that extracts disentangled feature abstractions at all levels to enhance the model's ability to generalize to new data from different hospitals/domains. Accordingly, the main contributions are:

\begin{itemize}
    \item The proposed approach, called ALFA, is an extension of the mDSDI \cite{bui2021exploiting} technique for DG. ALFA disentangles the components of SSL, domain-invariant, and domain-specific representations to reduce redundancy and improve generalization to unseen target data. By exploiting all levels of feature abstraction, ALFA strives to fully utilize the available information in the dataset.

    \item The mDSDI \cite{bui2021exploiting} approach utilizes adversarial training to extract domain-invariant features, but it can be unstable due to a non-differentiable step (gradient reversal). Therefore, a loss function called ``\emph{soft class-domain alignment}" is proposed to minimize the average divergence between two domain probability distributions and a target probability distribution representing a soft class label for each class. This loss function provides better stability during optimization and more distinct latent space for the representation.
    
    \item To evaluate the effectiveness of the proposed improvement, we conduct experiments on two public datasets: the PACS \cite{li2017deeper} benchmark for DG and a Renal Cell Carcinoma (RCC) subtyping task extracted Whole Slide Images (WSI) of TCGA \cite{TCGA} data portal. 
    
\end{itemize}
\section{Methods}

\begin{figure*}
\begin{minipage}{.6\textwidth}
  \centering
  \begin{overpic}[width=\textwidth,clip, trim=4cm 14.5cm 0cm 4cm]{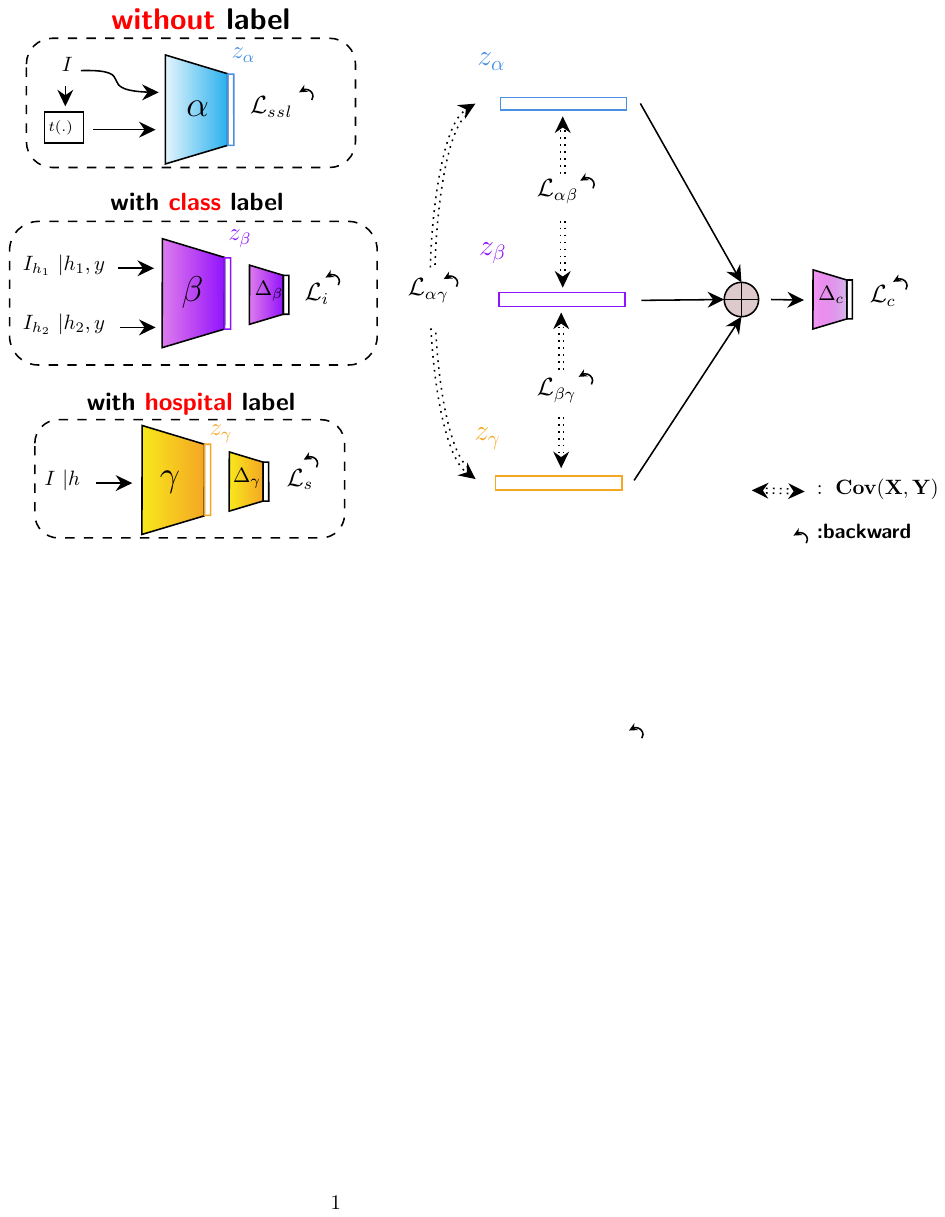}
    \put (40,52) {``Phase I"}
  \end{overpic}
  \label{fig:graphical1}
\end{minipage}%
\begin{minipage}{.3\textwidth}
  \centering
  \begin{overpic}[width=\textwidth,clip, trim=5cm 17cm 6cm 4cm]{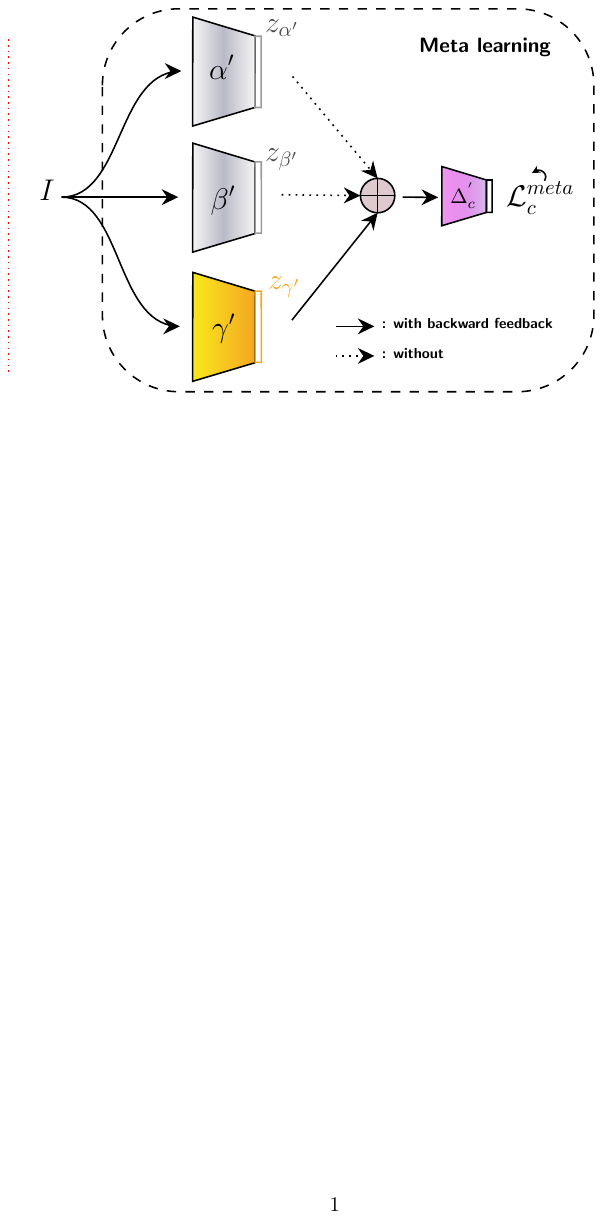}
    \put (40,82) {``Phase II"}
  \end{overpic}
  \label{fig:graphical2}
\end{minipage}
\caption{ALFA has two phases: In Phase I, three feature extractors extract different levels of feature abstraction, and disentangled features are concatenated for classification. In Phase II, updated feature extractors' representations are concatenated and fed into the updated classifier to update parameters in a Meta-learning fashion while $\alpha'$ and $\beta'$ feature extractors remain frozen.}
\label{fig:graphical}
\end{figure*}

The key concepts underlying the invented approach have been discussed up to this point. In this section, the specific details of how ALFA is implemented will be delved into. A visual representation and a high-level overview of ALFA framework are provided in Fig. \ref{fig:graphical}. It should be noted that the symbols ``$\oplus$" and ``$\circ$" have been employed as concatenation and union operators, respectively.

Several components are included in the invented integrated network: (1) an SSL representation $z_\alpha^{I} = \alpha(I;\theta_\alpha)$), with $\alpha$ being the SSL encoder that is parameterized by $\theta_\alpha$; (2) a domain-invariant representation $z_\beta^{I} = \beta(I;\theta_\beta)$), parameterized by $\theta_\beta$, which serves as the domain invariant feature extractor; (3) domain-specific representation $z_\gamma^{I} = \gamma(I;\theta_\gamma)$), where $\gamma$ stands for the domain-specific feature extractor parameterized by $\theta_\gamma$; (4) a domain aligner, parameterized by $\theta_{\Delta_\beta}$, denoted as $\Delta_\beta(z_\beta^I; \theta_{\Delta_\beta}): z_\beta^I \rightarrow \overline{1:N_c} $; (5) a domain classifier, parameterized by $\theta_{\Delta_\gamma}$, represented as $\Delta_\gamma(z_\gamma^I; \theta_{\Delta_\gamma}): z_\gamma^I \rightarrow \overline{1:N_{h}}$, where $N_c$ and $N_h$ refer to the number of classes, and the number of participating hospitals/domains in the training, respectively; (6) a regular classifier, parameterized by $\theta_c$, i.e.,  $\Delta_c(z_\alpha^I \oplus z_\beta^I \oplus z_\gamma^I; \theta_c): z_\alpha^I \oplus z_\beta^I \oplus z_\gamma^I \rightarrow \overline{1:N_{c}}$. The hospital and images sample spaces are represented by $\mathcal{H}$ and $\mathcal{I}$ respectively. Images, or $I$, are denoted by their target labels, or $y$, and hospital labels, or $h$, as $(I,y,h)$.

\subsection{Phase I: Extracting different levels of feature abstraction}
Phase I consists of multiple steps: Histopathology-tailored self-supervised learning, determining general features across hospitals, finding hospital-specific features, feature pair disentanglement, establishing the classification loss, and finally inference and training.  
\subsubsection{Construction of Histopathology-tailored Self-supervision Representations: Capturing the ``Synthetic-Invariant Features"}

At the core of ``Phase I" is the training of an SSL feature extractor, represented by $\alpha(I; \theta_\alpha)$. This component is designed to discern and learn generalizable features, which will become instrumental for accomplishing downstream tasks. The learning process employs a technique known as a triplet loss, a methodology similar to the one detailed in the study by Wang et al. \cite{wang2015unsupervised}.

The first step in this learning process involves the creation of pseudo-classes. These are generated by applying a set of transformations, denoted as $\mathcal{T}$, to a single image. This leads to the production of an augmented image, symbolized as $I_t = t(I)|t \sim \mathcal{T}$.

An image originating from a different pseudo-class, denoted as $I_d$, is also utilized in the training process. Together, these images serve as training inputs for the network, which employs the following triplet loss formula:
\begin{align}
   \mathcal{L}_\text{SSL} = \max(||z_\alpha^{I} - z_\alpha^{I_t}||_2 - ||z_\alpha^{I} -z_\alpha^{I_d}||_2 + M, 0),
   \label{eq:l_triplet}
\end{align}

In this equation, $||.||_2$ signifies the Euclidean distance and $M$ represents the margin, which is set heuristically at 1.5. This formula guides the network's learning process by maximizing the intra-class similarity and minimizing the inter-class similarity. 
The distribution shift is not known a priori but the most prevalent shift according to Fig. \ref{fig:differences} are color/stain or shape or resolution differences. Accordingly, transformations set $\mathcal{T}$ includes HED jitter (a histopathology tailored jitter that works on HED channel instead of RGB channels) with a jitter parameter of $\theta = 0.05$, as prescribed by Tellez et al., 2018 \cite{tellez2018whole}. Additionally, random affine transformations are applied, including rotation (varying between -10 and 10 degrees), translation (ranging from 0 to 0.1), and shear (from -1 to 1) in both $x$ and $y$ directions. For the resolution differences, we used resizing images to a lower resolution, then resizing them back to their original size, effectively achieving pixelation. These transformations contribute to the generation of diverse pseudo-classes, providing a wide range of inputs to strengthen the learning process.
In Eq. \ref{eq:l_triplet}, the margin, represented by $M$, is set to 1.5. The mining strategy employed is semi-hard mining, for which the margin is specified as 0.7. 

\subsubsection{Hospital-invariant representations: ``general features across hospitals"}

Aligning class relationships across hospitals promotes more transferable knowledge for model generalization compared to individual hard label prediction \cite{dou2019domain}. This study aims to impose an overall pattern of retrieved features that represent the intrinsic similarity between the semantic structures of different classes. Soft labels and class labels are used for consistency across domains.

\noindent
\textbf{-- Soft Confusion Matrix:} 
Let $\overline{z}_c^{(k)}$ denote the mean of class $c$ in the domain $k$ in the embedding space. We use \emph{softmax} activation for representing the probability of belonging to classes as $s_c^{(k)} = \mathit{softmax}(\Delta_\beta(\overline{z}_c^{(k)})/\tau)$, where $\tau > 1$ is the temperature.
The group of soft labels, $[s_c^{(k)}]_{c=1}^C $, serves as a form of \emph{soft confusion matrix} associated with a particular domain/hospital. By combining the class labels with this soft confusion matrix, our goals, i.e., domain alignment and retaining the class relationship, can be fulfilled. 

\noindent
\textbf{-- Soft Class Label Injection:} In addition to this soft confusion matrix, proposed in \cite{dou2019domain}, we have injected the soft class labels through a discrete probability density function that represents each class as follows
\begin{align}
p_c := \{i|i=\delta_i^{c}+(1-\delta_i^{c})(\frac{\zeta}{n_c-1})\},
\label{eq:pc}
\end{align}

where $\delta_i^{c}$ is the Kronecker delta which is defined as $\delta_i^{c} := \left\{\begin{matrix}
1 & \text{if} ~~i=c\\ 
0 & ~~\text{otherwise}
\end{matrix}\right.$, and $\zeta \lessapprox 1$ is a constant value indicating a high probability value (e.g., $\zeta = 0.9$), and $n_c$ is the number of classes.

Overall, the aim is to minimize the average divergence \cite{sgarro1981informational} over all the $C$ classes between three distributions: two arbitrary hospitals/domains samples drawn from the two training hospitals' images, $(I_{h_1},.,h_1)\sim \mathcal{H}_{1}, (I_{h_2},.,h_2)\sim \mathcal{H}_{2}$, and the probability distribution of each class, defined in Eq. \ref{eq:pc}. 

The \emph{Soft Class-Domain Alignment} loss which serves as the domain-invariant loss in our design is defined as 

\begin{equation}
\begin{aligned}
&\mathcal{L}_\text{i}((I_{h_1},.,h_1), (I_{h_2},.,h_2); \theta_\beta \circ \theta_{\Delta_{\beta}}) := \\
& ~~~~\frac{1}{C} \sum_{c=1}^C \frac{1}{6} \Big[ D_\text{KL} (s_c^{(h_{1})} \| s_c^{(h_{2})}) + D_\text{KL}(s_c^{(h_{2})} \| s_c^{(h_{1})}) +  \\
&~~~~~~~~~~~~~~~~~~~~D_\text{KL} (p_c \| s_c^{(h_{2})}) + D_\text{KL}(s_c^{(h_{2})} \| p_c)+ \\
&~~~~~~~~~~~~~~~~~~~~ D_\text{KL} ( s_c^{(h_{1})} \| p_c) +
D_\text{KL}(p_c \| s_c^{(h_{1})}) \Big], \label{equation_l_gen}
\end{aligned}
\end{equation}

where $D_\text{KL}(p \| q) = \sum_{r} p_r \log \frac{p_r}{q_r}$, and $\theta_\beta \circ \theta_{\Delta_{\beta}}$ is the union of all the parameters for $\beta$ feature extractor and $\Delta_\beta$ classifier.

The inclusion of our histopathology-tailored SSL features as a level of feature abstraction could raise the question of how distinct SSL features are from the hospital-invariant features in our design. The subsequent theorem and lemma aim to provide clarity on this matter:

\begin{theorem}
Given transformations $T_1$ for $\alpha$ (synthetic distribution shifts assuming in the self-supervision pretext task) and $T_2$ for differences of hospitals (due to the real distribution shift across domains/ hospitals), and an optimization objective to minimize the covariance between $z_{\alpha}$ and $z_\beta$ obtained respectively by $T_1$ (explicit data augmentation) and $T_2$ (implicit changes due to sources of distribution shifts), are distinct and uncorrelated in the feature space and both contribute unique information for mapping from feature space to label space.
\end{theorem}

\begin{lemma}
The following assertions hold for the features $z_\alpha = \alpha(T_1(I))$ and $z_\beta = \beta(T_2(I))$ for a given image $I$:
\begin{enumerate}
\item $z_\alpha$ and $z_\beta$ are uncorrelated:

If the optimization objective successfully minimizes the covariance, the covariance between $z_\alpha$ and $z_\beta$, $\text{Cov}(z_\alpha, z_\beta)$, will be close to zero. This indicates that $z_\alpha$ and $z_\beta$ are uncorrelated, i.e., changes in $z_\alpha$ do not predict changes in $z_\beta$ and vice versa.

\item $z_\alpha$ and $z_\beta$ contribute unique information to the mapping:

Let us consider a mapping function $M$ that maps the feature space to the label space. For $z_\alpha$ and $z_\beta$, the mapping function $M$ can be written as $M(z_\alpha, z_\beta)$. If $z_\alpha$ and $z_\beta$ are uncorrelated, removing one from the mapping will reduce the information provided by $M$. That is, $M(z_\alpha, \emptyset) \neq M(z_\alpha, z_\beta)$ and $M(\emptyset, z_\beta) \neq M(z_\alpha, z_\beta)$.

Therefore, given transformations $T_1$ for $\alpha$ and $T_2$ for domain shifts, and an optimization objective to minimize the covariance between the resulting augmentation-based self-supervised features and invariant features are distinct and uncorrelated in the feature space and both contribute unique information for mapping from feature space to label space.
\end{enumerate}
\end{lemma}

\subsubsection{Hospital-specific representations: ``least general or specific features"}
To extract the most specific features, similar to \cite{bui2021exploiting}, $\gamma(I;\theta_\gamma)$ is used for feature extraction followed by the $\Delta_{\gamma}$ domain classifier that is trained in a supervised manner using cross-entropy loss to predict the domain/hospital label:
\begin{align}
    \mathcal{L}_\text{s} := -\mathbb{E}_{(I,.,h) \sim \mathcal{I}} h \log{\Delta_{\gamma}(z_\gamma^{I};\theta_{\Delta_{\gamma}})}.
\end{align}

\begin{figure*}[!t]
    \centering
    
    \includegraphics[clip, trim=0cm 0cm 0cm 0cm, width=0.8\textwidth]{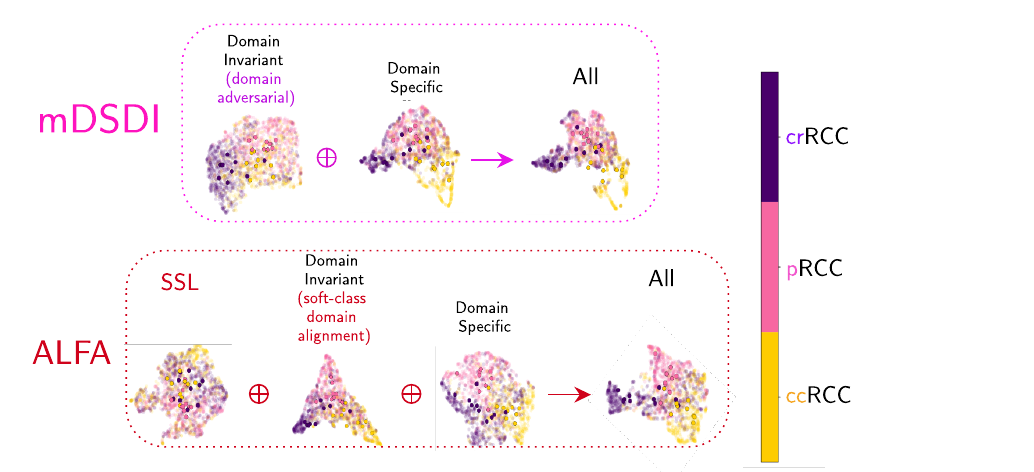}
    
    \caption{2D feature embeddings for the feature extractors in mDSDI \cite{bui2021exploiting} versus in ALFA: (target hospital: `NCI'). `All' is the concatenation of domain-specific and domain-invariant representations for the mDSDI \cite{bui2021exploiting} (up), and SSL, domain-invariant, and domain-specific representations for ALFA (bottom). Opaque-shaded scatters are WSIs representations obtained by averaging on patches' representations (transparent-shaded).}
    \label{fig:embeddings_rcc}
\end{figure*}

\subsubsection{Disentanglement loss between pairs of extracted features}
To prevent redundancy and ensure diversity in our feature extractors, we need to disentangle their resulting representations from each other. This can be achieved by zeroing the covariance matrix between pairs of random vectors, such as $z_a$ and $z_b$. A zero-covariance matrix indicates that the variables are independent and have no correlation or effect on each other. To enforce this disentanglement, we define pairwise covariance loss functions between each pair of $\alpha$, $\beta$, and $\gamma$ feature extractors' representations as 
\begin{align}
&\mathcal{L}_{\alpha\beta} :=  -\mathbb{E}_{I \sim \mathcal{I}} \big[||\operatorname{Cov}(z_{\alpha}^I, z_{\beta}^I)||_2\big], \\
& \mathcal{L}_{\alpha\gamma} :=  -\mathbb{E}_{I \sim \mathcal{I}} \big[||\operatorname{Cov}(z_{\alpha}^I, z_{\gamma}^I)||_2\big], \\
& \mathcal{L}_{\beta\gamma} :=  -\mathbb{E}_{I \sim \mathcal{I}} \big[||\operatorname{Cov}(z_{\beta}^I, z_{\gamma}^I)||_2\big].
\end{align}

\subsubsection{Classification loss using aggregation of extracted features}

The goal of the classifier $\Delta_c(z_\alpha^I \oplus z_\beta^I \oplus z_\gamma^I; \theta_c): z_\alpha^I \oplus z_\beta^I \oplus z_\gamma^I \rightarrow \overline{1:N_{c}}$ is to classify the images according to their hard class labels using a concatenation of all the extracted features:
\begin{align}
    \mathcal{L}_\text{c} := -\mathbb{E}_{(I,y,.) \sim \mathcal{I}} \left[y \log{\Delta_{c}(z_\alpha^I \oplus z_\beta^I \oplus z_\gamma^I; \theta_c)}\right],
\end{align}
where $y$ is the target class label for image $I$.

\subsubsection{Inference and training in Phase I --}
Using all loss functions, the feature extractors and modules are updated via
\begin{align}
    \mathcal{L}_\text{total} := a_1\mathcal{L}_{\text{SSL}} + a_2\mathcal{L}_i + a_3\mathcal{L}_s + \\ 
    \nonumber a_4\mathcal{L}_{\alpha\beta} + a_5\mathcal{L}_{\alpha\gamma} + a_6\mathcal{L}_{\beta\gamma} + a_7\mathcal{L}_c,
    \label{eq:l_total}
\end{align}

\noindent where $a_i$ coefficients are selected as balanced parameters between loss functions and all were set to 1.0 in our experiments as the loss values were in the same range. Through backward the total loss, i.e., $\mathcal{L}_\text{total}$, the updated encoders, i.e., $\alpha'(I;\theta_{\alpha'})$, $\beta'(I;\theta_{\beta'})$, $\gamma'(I;\theta_{\gamma'})$, and classifiers $\Delta_\beta'(I;\theta_{\Delta_\beta'})$, $\Delta_\gamma'(I;\theta_{\Delta_\gamma'})$ and $\Delta_c'(I;\theta_{\Delta_c'})$ are obtained.

\subsection{Phase II: Meta-learning for generalization improvement}

To adapt the domain-specific representation $z_\gamma$ to the target domain using information from source domains, we use the same meta-learning framework as \cite{bui2021exploiting}. The $\alpha'$ and $\beta'$ feature extractors remain frozen while the $\gamma'$ feature extractor and $\Delta_c'$ classifier are updated. We aim to update $\omega=\theta_{\gamma'} \circ \theta_{\Delta_c'}$ through meta-learning by dividing each hospital data $\mathcal{H}_k$ into disjoint meta-train $\mathcal{H}_{k}^{tr}$ and meta-test $\mathcal{H}_{k}^{te}$ sets and the objective is to
\begin{align}
    \min_{\omega}{~~~~ \mathcal{L}_c^\text{meta} := f(\omega - \nabla f(\omega, \mathcal{H}_{k}^{tr}), \mathcal{H}_{k}^{te})},
\end{align}
where
\begin{align}
f(\omega= \theta_{\gamma'} \circ \theta_{\Delta_c'}, \mathcal{H}_k) = \\
    \nonumber
    ~~ -\mathbb{E}_{(I_k,y_k,k) \sim \mathcal{H}_k} \big[ \\
    \nonumber
    ~~ y_k \log{\Delta_{c}'(z_{\alpha'}^{I_k} \oplus z_{\beta'}^{I_k} \oplus \gamma'(I_k, \theta_{\gamma'}); \theta_{\Delta_c'})}\\
    \nonumber
    \big],
\end{align}
where $y_k$ and $k$ are the target class label and hospital label, respectively, for image $I_k$.

\begin{figure*}[!t]
    \centering
    \resizebox{.8\textwidth}{!}{
    \includegraphics[clip, trim=0cm 0cm 0cm 0cm, width=0.8\textwidth]{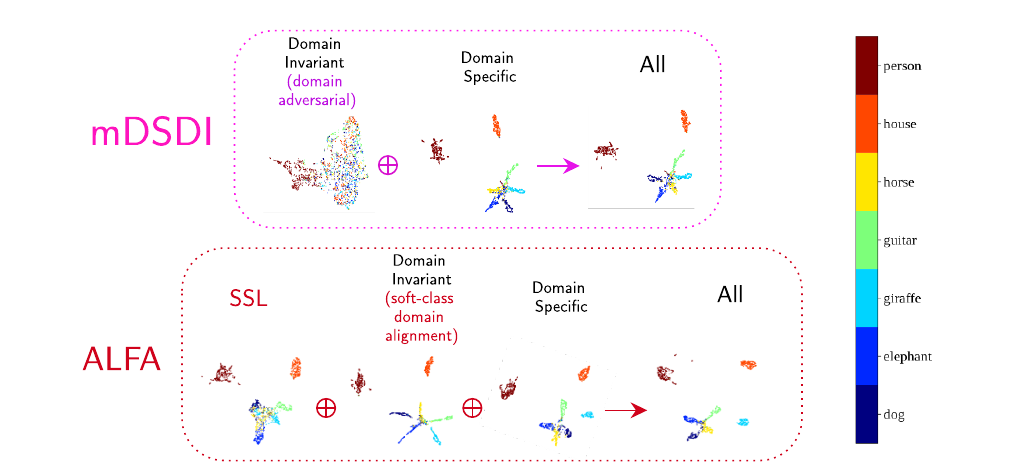}
    }
    \caption{2D feature embedding for the feature extractors in mDSDI \cite{bui2021exploiting} (upper row) versus in ALFA (ours) (bottom row), target domain:`Photo' on PACS.}
    \label{fig:embeddings_pacs}
\end{figure*}

\section{Experiments and Results}
The study evaluates the effectiveness of the proposed method, ALFA, against mDSDI \cite{bui2021exploiting}, HA \cite{sikaroudi2022hospital}, and ERM through a leave-one-\emph{domain/hospital}-out evaluation using data from multiple hospitals/domains. The evaluation includes reporting ``accuracy" for the target (hold-out) domain/hospital, as well as ``AUROC" and ``recall" metrics for RCC subtyping, which is important for cancer diagnosis tasks.

\subsection{Datasets}
\begin{figure}
    \centering
    \includegraphics[clip, trim=2cm 0.2cm 2cm 0cm, width=0.4\textwidth]{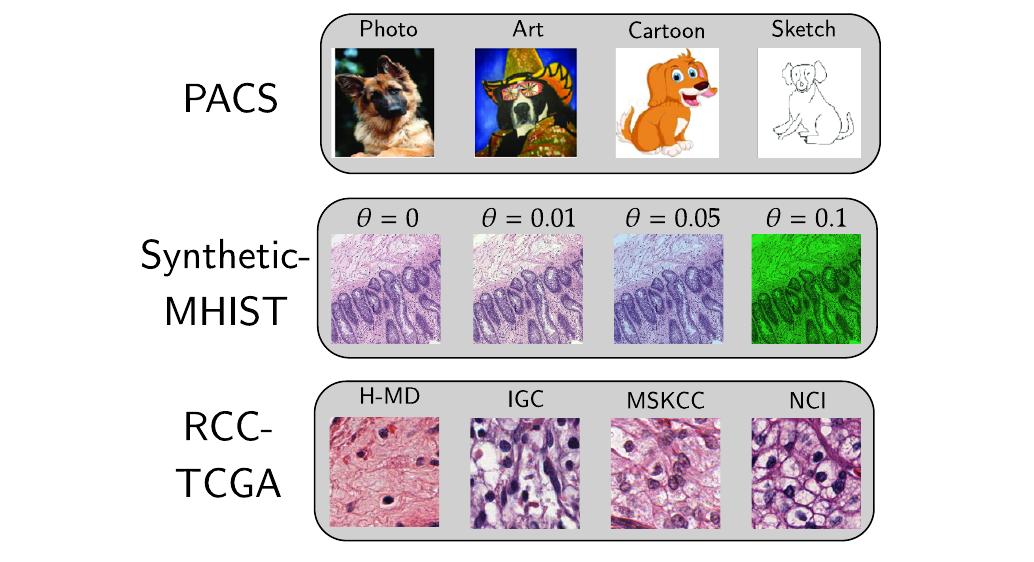}
    \caption{Different datasets and their domains in this study.}
    \label{fig:datasets}
\end{figure}

In this study, we evaluated our approach using three unique benchmarks: PACS, a Domain Generalization benchmark; synthetic-MHIST, a synthetic histopathology dataset mimicking staining variances, derived from the MHIST dataset by adding HED jitter; and a task involving Renal Cell Carcinoma (RCC) subtyping (Fig. \ref{fig:datasets}).

\noindent
\textbf{-- PACS:} P(hoto), A(rt), C(artoon), S(ketch) is a benchmark for DG on natural images. 
    It has been created by intersecting the classes in Caltech256 (Photo), Sketchy (Photo,Sketch) \cite{sangkloy2016sketchy}, TU-Berlin (Sketch) \cite{eitz2012humans} and Google Images (Art painting, Cartoon, Photo). 
    This benchmark includes four domains (Photo, Sketch, Cartoon, Painting), and 7 common categories ‘dog’, ‘elephant’, ‘giraffe’. ‘guitar’, ‘horse’, ‘house’, and ‘person’ with a total 9991 images.

\noindent
\textbf{-- Sythetic-MHIST}: MHIST \cite{wei2021learn} is a histopathology dataset containing two different class labels, i.e. Hyperplastic Polyp (HP), and Sessile Serrated Adenoma (SSA) with a total of 3152 images.  It involves the distinction between the clinically-important binary categories of HPs and SSAs, which is a challenging problem with considerable pathologists' inter-variability. In this study, we synthetically created four different domains by applying HED jitter borrowed from \cite{tellez2018whole} with different degrees to this tiny dataset. HED jitter is a histopathology-tailored color jitter that applies random distortions in the HED color space.
     

\noindent
\textbf{-- RCC subtyping dataset from TCGA:}
The RCC dataset \cite{sikaroudi2022hospital,hosseini2023proportionally}, comprises patches of various Renal Cell Carcinoma (RCC) cancer subtypes collected from five different hospitals. Due to the absence of certain cancer subtypes, two of the hospitals' data have been merged. The dataset comprises 4 image repositories: (1) H-MD, (2) MSKCC, (3) IGC, and (4) NCI from TCGA. The dataset contains $\approx 70k$ patches of size $224 \times 224$. 

\subsection{Experimental Setup}

The backbone of all feature extractors was the ResNet18 \cite{he2016deep}, pre-trained on ImageNet \cite{krizhevsky2017imagenet}, with all of its batch normalization layers frozen as per the guidelines given in \cite{seo2020learning}. All features were embedded to a size of 512. The Adam optimizer was employed with an initial learning rate of 5e-5. A batch size of 32 was established and set the maximum number of iterations to 3000.

\begin{figure*}[!h]
    \centering
    \includegraphics[clip, trim=2cm 3cm 2cm 3cm, width=0.7\textwidth]{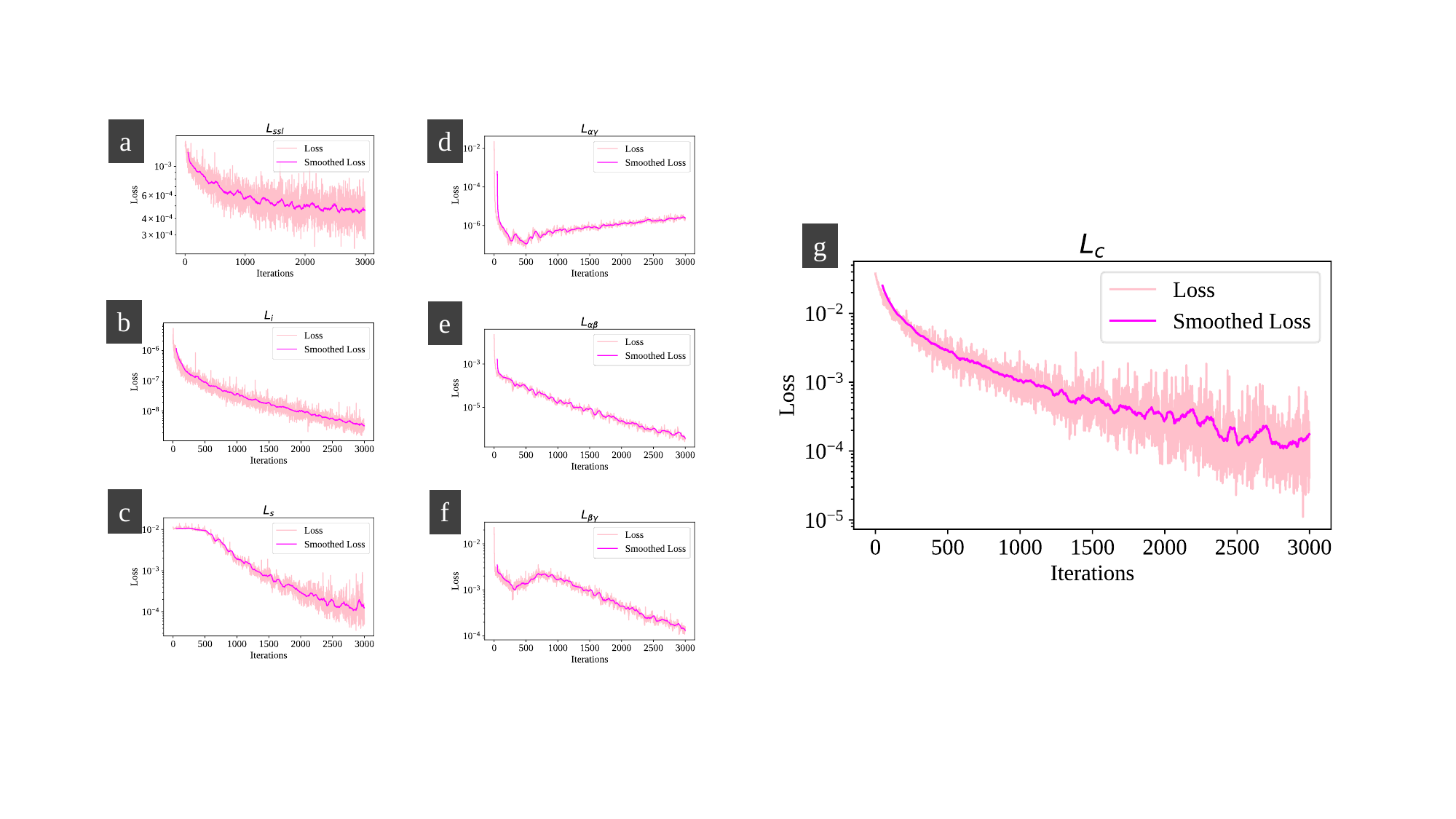}
    \caption{The behavior of ALFA's loss functions during the training when the hold-out set was Art on PACS. (a) $\mathcal{L}_\text{SSL}$, (b) $\mathcal{L}_i$, (c) $\mathcal{L}_s$, (d) $\mathcal{L}_{\alpha\gamma}$, (e) $\mathcal{L}_{\alpha\beta}$, (f) $\mathcal{L}_{\beta\gamma}$, (g) $\mathcal{L}_c$}
    \label{fig:embeddings_pacs}
\end{figure*}

\subsection{Results}
\noindent
\textbf{-- Losses convergence:} During the training, it was found that $\gamma$ feature extractor was dominating over other feature extractors and potentially causing a dampening effect on their contributions to the model's overall performance. Hence, we added layer normalization \cite{ba2016layer} whenever the representations are concatenated to address this issue. With this modification, all losses converged almost simultaneously according to Fig. \ref{fig:embeddings_pacs}.

\noindent
\textbf{-- Low-dimensional Embedding Visualization:} In accordance with the best practices suggested in the original UMAP paper \cite{mcinnes2018umap}, PCA \cite{wold1987principal} was applied to obtain the first 50 principal components, followed by UMAP \cite{mcinnes2018umap} for further dimensionality reduction to 2. According to Figs. \ref{fig:embeddings_pacs} and \ref{fig:embeddings_rcc}, ALFA's domain-invariant approach yields a more powerful discriminatory representation for different RCC subtypes or different categories on PACS, compared to mDSDI \cite{bui2021exploiting}, for domain-specific representations. In other words, ALFA's domain-invariant encoder learns some features that are also learned by mDSDI's domain-specific encoder. ALFA's SSL representation provides useful representation, which seems even better than mDSDI's domain-invariant features according to these figures.

\subsubsection{PACS dataset classification task}
The accuracy of mDSDI \cite{bui2021exploiting} and ALFA applied on PACS have been reported in Table \ref{tab:pacs}. It can be seen in Table \ref{tab:pacs}, except for the ‘Sketch’ with a high semantic shift in comparison to the rest of the target domains, ALFA outperforms mDSDI \cite{bui2021exploiting} with an average accuracy of $83.75\pm6.72\%$ compared to mDSDI's \cite{bui2021exploiting} average accuracy of $80.34\pm5.60\%$. According to this, ALFA cannot only be effective for generalization to unseen hospitals but it can also be effective for DG tasks for natural images.

\begin{table}[!h]
\centering
\caption{Results on PACS dataset}
\label{tab:pacs}
\resizebox{0.7\columnwidth}{!}{
\begin{tabular}{||ccccc||}
\hline
\multicolumn{2}{||c}{}        & \multicolumn{3}{c||}{Accuracy (\%)}

\\ \hline

Target       & Source       & ERM                  & mDSDI \cite{bui2021exploiting}                 & ALFA(ours)                 \\ \hline \hline
Photo        & \{A,C,S\}        & 91.98                & 90.06                 & \cellcolor{green!25}96.15                 \\ \hline
Art          & \{P,C,S\}        & 76.85                & 76.27                 & \cellcolor{green!25}83.10                 \\ \hline
Cartoon      & \{P,A,S\}        & 74.87                & 76.20                 & \cellcolor{green!25}78.71                 \\ \hline
Sketch       & \{P,A,C\}        & 76.76                & \cellcolor{green!25}78.85                 & 78.41                 \\ \hline
\multicolumn{2}{||c}{Average}           & $80.11	\pm6.76$           & $80.34\pm5.60$          & \cellcolor{green!25}$84.09 \pm 7.06$        \\ \hline
\end{tabular}
}
\end{table}

\subsubsection{Synthetic-MHIST classification task} Considering the results on the synthetic-MHIST dataset in Table \ref{tab:synthetic}, we can see that ALFA (ours) outperforms mDSDI \cite{bui2021exploiting} in all cases except for when the target domain is $\theta=0.05$. On average, ALFA achieves higher accuracies, with an average accuracy of $84.17\pm3.17\%$ compared to mDSDI's average accuracy of $82.65\pm5.02\%$. For the case when $\theta=0.5$ is the target domain, which represents the most challenging target domain with the highest degree of corruption and accordingly highest distribution shift in comparison to the rest of the source domains, the ALFA has a significant improvement over mDSDI and ERM. This can be attributed to the fact that incorporating self-supervision representations, as ALFA does, can lead to improved performance in generalization to unseen domains compared to using only specific and domain-invariant features, as mDSDI does.

\begin{table}[!h]
\centering
\caption{Results on Synthetic-MHIST dataset}
\label{tab:synthetic}

\resizebox{0.8\columnwidth}{!}{
\begin{tabular}{||ccccc||}
\hline
Target                             & Source                                                & ERM                  & mDSDI                 & ours                 \\ \hline\hline
\multicolumn{1}{||c}{$\theta=0$}   & \multicolumn{1}{c}{$\theta=\{0.01, 0.05, 0.5\}$}        & 84.65                & 85.16                  & \cellcolor{green!25}86.39              \\ \hline
\multicolumn{1}{||c}{$\theta=0.01$}& \multicolumn{1}{c}{$\theta=\{0,0.05,0.5\}$}           & 84.75                & 86.18                 & \cellcolor{green!25}86.28                 \\ \hline
\multicolumn{1}{||c}{$\theta=0.05$}& \multicolumn{1}{c}{$\theta=\{0,0.01,0.5\}$}           & 85.26                & \cellcolor{green!25}85.47                 & \cellcolor{green!25}85.47                 \\ \hline
\multicolumn{1}{||c}{$\theta=0.5$}& \multicolumn{1}{c}{$\theta=\{0, 0.01,0.05\}$}          & 75.64                & 73.80                 & \cellcolor{green!25}78.6                 \\ \hline
\multicolumn{2}{||c}{Average}           & $82.57\pm3.93$           & $82.65\pm5.02$          & \cellcolor{green!25}$84.17\pm3.17$        \\ \hline
\end{tabular}
}
\end{table}

\subsubsection{RCC subtyping classification task}
The accuracy of mDSDI \cite{bui2021exploiting}, HA \cite{sikaroudi2022hospital} approach, and ALFA applied on RCC subtyping task has been reported in Table \ref{tab:RCC}.

In the context of H-MD, ALFA (ours) exceeds the performance of both ERM and mDSDI, achieving an accuracy of 65.52\% as opposed to 72.49\% and 51.72\% respectively. However, it falls short when compared to the HA \cite{sikaroudi2022hospital} method, which reaches 75.29\% accuracy. This outcome may indicate that, considering H-MD encompasses two distinct data sources, methods that focus on extracting invariant features are more effective. In this regard, HA \cite{sikaroudi2022hospital}, which specifically targets hospital-invariant features, outperforms both ALFA and mDSDI.

\begin{table*}[!t]
\centering
\footnotesize
\caption{Results on RCC subtyping task}
\label{tab:RCC}
\resizebox{0.8\textwidth}{!}{%
\begin{tabular}{||c|p{1.5cm}|cccc|cccc|cccc||}
\hline
\multicolumn{2}{||c}{}        & \multicolumn{4}{c}{Accuracy (\%)}                                          & \multicolumn{4}{c}{AUROC (\%)}                                        & \multicolumn{4}{c||}{Recall (\%)}                                       \\ \hline
Target  & Source            & ERM   & mDSDI \cite{bui2021exploiting} & HA \cite{sikaroudi2022hospital} & ALFA           & ERM  & mDSDI \cite{bui2021exploiting} & HA \cite{sikaroudi2022hospital} & ALFA           & ERM  & mDSDI \cite{bui2021exploiting} & HA \cite{sikaroudi2022hospital} & ALFA           \\ \hline  \hline
IGC     & \{NCI, MSKCC, H-MD\} & 75.86 & 86.20 & 70.42 & \cellcolor{green!25}86.21 & 93.23 & \cellcolor{green!25}95.78 & 88.36 & 95.33 & 57.14 & 82.88 & 62.38 & \cellcolor{green!25}85.39 \\ \hline
NCI     & \{IGC, MSKCC, H-MD\} & 81.82 & 72.73 & 83.38 & \cellcolor{green!25}86.36 & 96.49 & 94.46 & 97.32 & \cellcolor{green!25}97.83 & 83.08 & 71.46 & 85.48 & \cellcolor{green!25}86.41 \\ \hline
MSKCC  & \{IGC, NCI, H-MD\}   & 86.73 & 85.71 & \cellcolor{green!25}88.19 & 84.69 & 95.91 & 95.89 & \cellcolor{green!25}96.47 &  95.99 & 82.99 & 87.05 & 85.32 & \cellcolor{green!25} 87.99 \\ \hline
H-MD   & \{IGC, NCI, MSKCC\}   & 72.49 & 51.72 & \cellcolor{green!25}75.29 & 65.52 & 85.38 & 88.37 & 90.16 & \cellcolor{green!25} 90.48 & 72.96 & 51.85 & \cellcolor{green!25}78.42 & 66.67 \\ \hline
\multicolumn{2}{||c|}{Average} & 
\specialcell{$79.22$ \\ $\pm 5.36$} & 
\specialcell{$74.09$ \\ $\pm 13.72$} & 
\specialcell{$79.32$ \\ $\pm 6.49$} &
\cellcolor{green!25}\specialcell{$80.69$ \\ $\pm 8.61$} & 
\specialcell{$92.75$ \\ $\pm 4.34$} & 
\specialcell{$93.62$ \\ $\pm 3.02$} & 
\specialcell{$93.08$ \\ $\pm 3.34$} &
\cellcolor{green!25}\specialcell{$94.90$ \\ $\pm 2.66$} & 
\specialcell{$74.07$ \\ $\pm 10.38$} &
\specialcell{$73.31$ \\ $\pm 13.36$} & 
\specialcell{$77.90$ \\ $\pm 8.44$} &
\cellcolor{green!25}\specialcell{$81.62$ \\ $\pm 8.50$} \\ \hline
\end{tabular}
}
\end{table*}

For the IGC dataset, both ALFA and mDSDI attain the same accuracy of 86.21\%, outperforming ERM and HA \cite{sikaroudi2022hospital}, which scored 75.86\% and 70.42\%, respectively. Given that both ALFA and mDSDI leverage a combination of invariant and hospital-specific features, it's reasonable to speculate that the inclusion of specific features has enhanced their overall performance in this context. In essence, the unique features sourced from the different hospitals appear to have provided valuable information that contributed to improved performance on the IGC dataset.

At the NCI, ALFA again leads with an accuracy of 86.36\%, outperforming ERM (81.82\%), mDSDI \cite{bui2021exploiting} (72.73\%), and HA \cite{sikaroudi2022hospital} (83.38\%).

In the case of MSKCC, ALFA's performance is notably lower with an accuracy of 84.69\%, when compared to both ERM (86.73\%) and HA \cite{sikaroudi2022hospital} (88.19\%). However, ALFA does surpass mDSDI, which has an accuracy of 85.71\%. This result might suggest that invariant features are more conducive to effective generalization in this context than specific features. Therefore, both mDSDI and ALFA, which rely on a combination of site-specific and site-invariant features, may not perform as well as HA \cite{sikaroudi2022hospital} or even ERM.

According to this table, ALFA outperforms mDSDI \cite{bui2021exploiting}, HA \cite{sikaroudi2022hospital}, and ERM with average accuracy, AUROC, and recall of $80.69\pm8.61\%$, $94.90\pm2.66\%$, and $81.62\pm8.50\%$, respectively. ALFA performed similarly to mDSDI and ERM in terms of AUROC and accuracy, but slightly better overall. However, ALFA significantly outperformed other methods in terms of recall metric, especially for the ``NCI", and ``IGC"  target hospitals.

\subsubsection{Ablation study}

In the conducted ablation study, each element of ALFA demonstrated its efficacy in improving overall generalization, as illustrated in Table \ref{tab:ablation}. We assessed feature extractors, labeled as $\alpha$, $\beta$, and $\gamma$, in various active or inactive states. Performance was evaluated across four subsets from PACS and RCC, with an average accuracy score calculated. When only $\alpha$ was active, the average accuracy was lowest, at $48.63\pm14.98$ for PACS and $46.50\pm12.20$ for RCC. There was a substantial increase in average accuracy to around $78.88\pm 6.99$ for PACS and similar for RCC, when either $\beta$ or $\gamma$ was activated while the others were inactive. The highest average accuracies, $84.09\pm7.06$ for PACS and $80.70\pm8.99$ for RCC, were achieved when multiple feature extractors were activated concurrently, highlighting the effectiveness of the complete ALFA method.

\section{Conclusions}

In this study, we introduced an inventive methodology named ALFA (All Levels of Features Abstraction), which exploits multiple feature extractors to disassociate features at disparate levels of abstraction. The efficiency of ALFA is substantiated across a range of benchmarks, such as the commonly utilized DG benchmark PACS, a synthetic-MHIST dataset, and Renal Cell Carcinoma subtype identification utilizing the TCGA database. ALFA successfully surpassed the existing state-of-the-art mDSDI approach in performance on PACS, underscoring its superior capabilities. Moreover, in comparison to mDSDI and ERM, ALFA showcased exceptional capability to generalize to previously unseen hospitals. In our approach, we employed a straightforward augmentation-based self-supervision technique that was tailored for histopathology, enabling the acquisition of invariant features. Concurrently, we introduced a novel domain alignment loss function, termed "soft class-domain alignment" loss. This technique skillfully extracts domain-invariant features, aligning data from two disparate sources with soft class labels serving as the reference distribution, thereby enhancing our design significantly. As advancements continue in the realm of self-supervision, the incorporation of more effective SSL features could potentially augment our design even further.

\begin{table}[!h]
\caption{Ablation on PACS and RCC: {Active feature extractor(s) is/are \textcolor{blue}{blue}, Deactivated one(s): \textcolor{gray}{gray}}}
\label{tab:ablation}
\centering
\resizebox{0.8\columnwidth}{!}{

\begin{tabular}{||c|cccc|c||}
\hline
            \textbf{PACS }& Photo & Art   & Cartoon & Sketch & Average         \\
            \hline
            \hline
\textcolor{blue}{$\alpha$}, \textcolor{gray}{$\beta$}, \textcolor{gray}{$\gamma$}       & 69.70 & 53.17 & 44.28   & 27.38  & $48.63 \pm14.98$ \\ \hline
\textcolor{gray}{$\alpha$}, \textcolor{blue}{$\beta$}, \textcolor{gray}{$\gamma$}        & 91.13 & 73.33 & 75.68   & 75.38  & $78.88 \pm 6.99 $  \\ \hline
\textcolor{gray}{$\alpha$}, \textcolor{gray}{$\beta$}, \textcolor{blue}{$\gamma$}       & 94.43 & 77.34 & 72.05   & 71.46  & $78.82\pm9.11$ \\ \hline
\textcolor{blue}{$\alpha$}, \textcolor{blue}{$\beta$}, \textcolor{gray}{$\gamma$}  & 90.11 & 75.48 & 76.23   & 75.99  & $79.45\pm6.03$  \\ \hline
\textcolor{blue}{$\alpha$}, \textcolor{gray}{$\beta$}, \textcolor{blue}{$\gamma$} & 94.73 & 78.61 & 73.93   & 72.02  & $79.82\pm8.75$   \\ \hline
\textcolor{gray}{$\alpha$}, \textcolor{blue}{$\beta$}, \textcolor{blue}{$\gamma$}  & 95.38 & 82.03 & 78.37   & 78.67  & $83.61\pm6.80$  \\ \hline 
\textcolor{blue}{ALFA} & 96.15 & 83.10 & 78.71   & 78.41  & $84.09\pm7.06$  \\ \hline \hline
            \textbf{RCC} & IGC & NCI & MSKCC & H-MD & Average\\
            \hline
            \hline
\textcolor{blue}{$\alpha$}, \textcolor{gray}{$\beta$}, \textcolor{gray}{$\gamma$}       & 62.34 & 48.23 & 40.12   & 35.30  & $46.50 \pm12.20$ \\ \hline
\textcolor{gray}{$\alpha$}, \textcolor{blue}{$\beta$}, \textcolor{gray}{$\gamma$}        & 77.43 & 62.34 & 64.28   & 64.08  & $67.03 \pm 6.28 $  \\ \hline
\textcolor{gray}{$\alpha$}, \textcolor{gray}{$\beta$}, \textcolor{blue}{$\gamma$}       & 80.34 & 65.78 & 61.38   & 61.10  & $67.15\pm8.11$ \\ \hline
\textcolor{blue}{$\alpha$}, \textcolor{blue}{$\beta$}, \textcolor{gray}{$\gamma$}  & 76.56 & 63.84 & 64.21   & 64.11  & $67.18\pm5.99$  \\ \hline
\textcolor{blue}{$\alpha$}, \textcolor{gray}{$\beta$}, \textcolor{blue}{$\gamma$} & 80.23 & 66.41 & 62.43   & 60.98  & $67.51\pm7.85$   \\ \hline
\textcolor{gray}{$\alpha$}, \textcolor{blue}{$\beta$}, \textcolor{blue}{$\gamma$}  & 80.77 & 72.43 & 73.11   & 64.42  & $72.68\pm5.67$  \\ \hline
\textcolor{blue}{ALFA} & 86.21 & 86.36 & 84.69   & 65.52  & $80.70\pm8.99$  \\ \hline
\end{tabular}
}
\end{table}

{\small
\bibliographystyle{ieee_fullname}
\bibliography{egbib}
}

\end{document}